\documentclass[conference]{IEEEtran}
\IEEEoverridecommandlockouts
\usepackage{hyperref}
\usepackage{graphicx}
\usepackage{xcolor,colortbl}
\usepackage{subcaption}
\usepackage{cite}
\usepackage{amsmath,amssymb,amsfonts}
\usepackage{algorithm}
\usepackage{algorithmic}
\usepackage{graphicx}
\usepackage{textcomp}
\usepackage{xcolor}
\usepackage{multirow}
\renewcommand{\algorithmicrequire}{\textbf{Input:}}
 \renewcommand{\algorithmicensure}{\textbf{Output:}}
\newcommand{\our}{\emph{FairBranch}}

\newcommand{\nt}{\emph{negative transfer}}
\newcommand{\ut}{\emph{bias transfer}}

\DeclareMathOperator*{\argmax}{argmax}

\def\BibTeX{{\rm B\kern-.05em{\sc i\kern-.025em b}\kern-.08em
    T\kern-.1667em\lower.7ex\hbox{E}\kern-.125emX}}
\begin{document}

\title{FairBranch: Mitigating Bias Transfer in Fair Multi-task Learning}

\author{
\IEEEauthorblockN{Arjun Roy\IEEEauthorrefmark{2}\IEEEauthorrefmark{4}\\ \tt\footnotesize arjun.roy@unibw.de}
\and
\IEEEauthorblockN{Christos Koutlis\IEEEauthorrefmark{3}\\ \tt\footnotesize ckoutlis@iti.gr}
\and
\IEEEauthorblockN{Symeon Papadopoulos\IEEEauthorrefmark{3}\\ \tt\footnotesize papadop@iti.gr}
\and
\IEEEauthorblockN{Eirini Ntoutsi\IEEEauthorrefmark{4}\\ \tt\footnotesize eirini.ntoutsi@unibw.de}
\and
\centering
\IEEEauthorblockA{\hspace{1cm}\IEEEauthorrefmark{2}Dept. Math \& CSc., Free University of Berlin; \hspace{1cm} \IEEEauthorrefmark{3}Information Technologies Institute, CERTH 
}
\and
\centering
\IEEEauthorblockA{
\hspace{7cm}\IEEEauthorrefmark{4}RI CODE, Bundeswehr University, Munich.
\thanks{Copyright © 2024 IEEE. International Joint Conference on Neural Networks (IJCNN). All rights reserved.}}
}

\maketitle

\begin{abstract}
The generalisation capacity of  Multi-Task Learning (\verb|MTL|) suffers when 
unrelated tasks negatively impact each other by updating shared parameters with conflicting gradients. This is known as \emph{negative transfer} and leads to a drop in \verb|MTL| accuracy compared to single-task learning (\verb|STL|). 
Lately, there has been a growing focus on the fairness of \verb|MTL| models, requiring the optimization of both accuracy and fairness for individual tasks. Analogously to negative transfer for accuracy, task-specific fairness considerations might adversely affect the fairness of other tasks when there is a conflict of fairness loss gradients between the jointly learned tasks - we refer to this as \ut. 
To address both negative- and bias-transfer in \verb|MTL|, we propose a novel method called \our{}, which branches the \verb|MTL| model by assessing the similarity of learned parameters, thereby grouping related tasks to alleviate negative transfer.  Moreover, it incorporates fairness loss gradient conflict correction between adjoining task-group branches to address bias transfer within these task groups. 
Our experiments on tabular and visual MTL problems show that \our{} outperforms state-of-the-art  \verb|MTL|s on both fairness and accuracy. Our code is available on \href{https://github.com/arjunroyihrpa/FairBranch}{github.com/arjunroyihrpa/FairBranch}
\end{abstract}

\begin{IEEEkeywords}
multitasking, fairness, negative-transfer, bias-transfer, task-grouping
\end{IEEEkeywords}

\section{Introduction}\label{sec.intro}
Multi-Task Learning (\verb|MTL|) traditionally involves deep neural networks trained with fully shared representation layers (parameters) common to all tasks followed by individual task-specific layers to improve model performance across multiple tasks~\cite{MTLsurvey22}. 
However, when tasks do not align in their optimisation directions, conflicting updates to the shared parameters may occur, i.e., they may attempt to update the shared parameters with gradients pulling in conflicting directions~\cite{yu2020surgery}, resulting in performance degradation of the \verb|MTL| model on specific tasks compared to \verb|STL| models~\cite{ruder2017overviewMTL}, a phenomenon commonly known as  \nt{} of knowledge \cite{stanfordICML20MTLG}. 

Lately, there has been a growing focus on the fairness of \verb|MTL| models~\cite{hu2023fairMTL,RoyNtoutsiECML22,MTLfairWang0BPCC21}, and it is shown that such models can make biased predictions for specific demographic groups characterized by a protected attribute, such as gender or race, across multiple tasks. 
Fair-MTL methods try to optimize for both accuracy and fairness~\cite{RoyNtoutsiECML22,MTLfairWang0BPCC21,hu2023fairMTL}, by incorporating, for example, a fairness loss alongside the accuracy loss for each task~\cite{2022fairnessreview,MTLfairWang0BPCC21}. 
Analogously to negative transfer for accuracy, bias transfer may occur in fair-MTL, where task-specific fairness considerations could negatively affect the fairness of other tasks, when conflicting fairness loss gradients emerge among jointly learned tasks.

\begin{figure*}
    \centering
     \begin{subfigure}[b]{0.47\linewidth}
         \centering
         \includegraphics[width=\columnwidth]{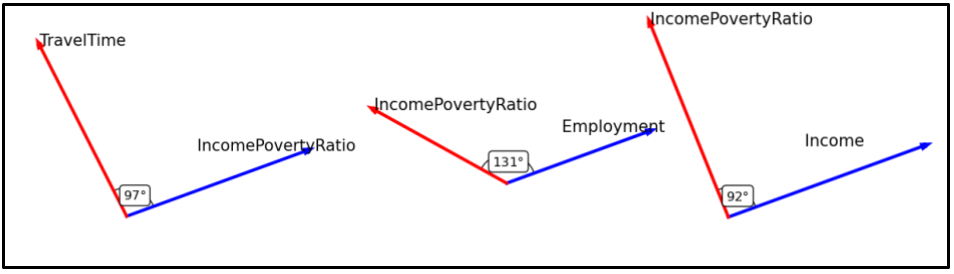}
         \caption{TAG}\label{fig.tag_ut}
    \end{subfigure}
    \begin{subfigure}[b]{0.47\linewidth}
         \centering
         \includegraphics[width=\columnwidth]{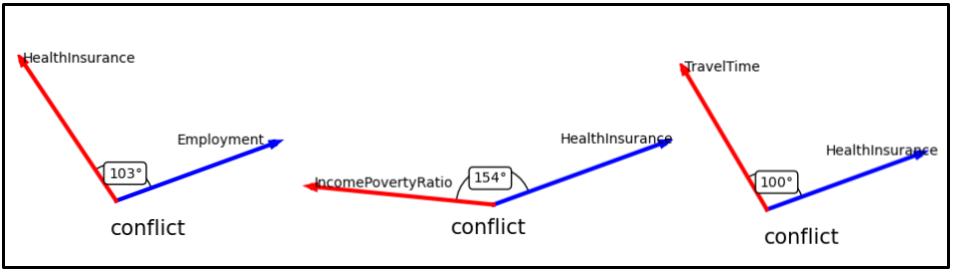}
         \caption{Recon}\label{fig.rec_ut}
    \end{subfigure}
    \caption{\small Fairness loss gradient conflicts observed in state-of-the-art MTLs addressing negative transfer of accuracy: (a) TAG~\cite{Google2021MTLG} using task-grouping and (b) Recon~\cite{guangyuan2022recon} using gradient correction on the ACS-PUMS Census Data 2018.}\label{fig.ex_fairconfs}
    \vspace{-0.1in}
\end{figure*}

In our paper, we aim to tackle the intertwined challenges of negative transfer and bias transfer in Multi-Task Learning (MTL). Negative transfer in vanilla MTL has been addressed through various methods, including balancing task-specific weights \cite{chen2018gradnorm,KDD19MTLGroup}, gradient conflict correction \cite{du2018gradient_similarity,guangyuan2022recon,yu2020surgery,wang2020gradvac}, employing branching model architectures \cite{lu2017FAFS,bruggemann2020btmas_branch,guangyuan2022recon}, and learning separate models for each task-group \cite{Google2021MTLG}. While using task-specific weights is cost-effective, determining them poses a significant challenge, especially when considering fairness-accuracy trade-offs for each task.
Moreover, methods solely relying on balancing task-weights, correcting gradients, or learning fixed task-group models are constrained by their fixed architecture \cite{guo2020mtlbranch}. Approaches addressing gradient conflicts can be computationally slow, as they necessitate computing and comparing conflicts for every possible task pair in each epoch, a challenge compounded in fair-MTL due to increased possibilities of conflict \cite{guangyuan2022recon}.
Notably, state-of-the-art methods in mitigating negative transfer fail to address fairness conflict issues, leading to bias transfer. In our experiments on the ACS-PUMS dataset (Fig.~\ref{fig.ex_fairconfs}), we illustrate the shortcomings of two prominent \verb|MTL| methods: TAG \cite{Google2021MTLG}, which employs task grouping, and Recon \cite{guangyuan2022recon}, which uses gradient correction. These results underscore the inability of negative transfer correction alone to resolve fairness conflicts.

Our proposed solution, \our{}, addresses both \nt{} and \ut{} by mitigating \nt{} through accuracy conflict-aware task grouping and countering \ut{} through fairness gradient conflict correction. We create task-group branches based on parameter similarity and correct fairness conflicts within each branch. This branching strategy helps mitigate accuracy loss gradient conflicts, as tasks with similar parameters exhibit similar loss gradient directions. By limiting fairness conflict correction to within task-group branches, our method scales effectively to a large number of tasks.

Our key contributions can be summarised as follows: i) We introduce the study of \ut{} (negative transfer of fairness) to enable bias-aware sharing of information among the tasks in MTL. ii) We propose \our{} \verb|MTL| that leverages parameter similarity to branch the network, and performs fairness loss gradient correction within each branch to mitigate bias transfer within the task groups. iii) We show that \our{}  outperforms state-of-the-art \verb|MTL| methods in addressing negative and bias transfer. 

\section{Related Works}\label{sec.related}
Related work can be categorized into two broad categories: MTL methods that tackle negative transfer (\nt{}) and fairness-aware MTL methods.

Various methods have been proposed to address negative transfer in vanilla MTL, including balancing task-specific weights \cite{chen2018gradnorm,KDD19MTLGroup}, gradient conflict correction \cite{du2018gradient_similarity,guangyuan2022recon,yu2020surgery,wang2020gradvac}, employing branching model architectures \cite{lu2017FAFS,bruggemann2020btmas_branch,guangyuan2022recon}, and learning separate models for each task-group \cite{Google2021MTLG}. Among these, methods that utilize task-grouping (e.g., TAG \cite{Google2021MTLG} and FAFS \cite{lu2017FAFS}) or are gradient conflict-aware (e.g., PCGrad \cite{kurin2022drawbcPCGRAD} and Recon \cite{guangyuan2022recon}) emerge as direct competitors to our approach. While task-grouping methods compare evaluated task loss output to compute groups, our approach groups tasks based on the learned parameter space, which we show is more effective in addressing the \nt{} problem. Our strategy for \nt{} is inspired by PCGrad but resolves conflicts only within task branches, requiring fewer conflict corrections and scaling better with a large number of tasks.

Fairness-aware MTL methods can be categorized into in-processing approaches like L2TFMT \cite{RoyNtoutsiECML22} and WB-fair \cite{MTLfairWang0BPCC21}, which modify the objective function by incorporating fairness losses alongside accuracy losses for each task, and post-processing approaches \cite{hu2023fairMTL} that learn data-driven distance estimators to adjust learned class boundaries. However, none of these prior works explicitly studied the problem of \nt{}. Our work falls under the in-processing category of fairness-aware learning, where we branch the model architecture based on parameter similarity in task-groups and then correct fairness conflicts within each task-group to address the joint problem of negative and bias transfer.

We provide a comparative overview of the various methods most relevant to our work in Table~\ref{tab:methods_comparison}. The evaluation dimensions include whether they address negative transfer, consider fairness, and incorporates dynamic architecture adaptation. Our method is the only one addressing all dimensions.
\begin{table}[htbp]
    \centering
    \caption{A comparative overview of SOTA}
    \label{tab:methods_comparison}
    \begin{tabular}{|c|c|c|c|}
        \hline
        \textbf{Methods} & \textbf{Negative Transfer} & \textbf{Fairness} & \textbf{Dynamic Architecture} \\
        \hline
        FAFS~\cite{lu2017FAFS} & \checkmark & - & \checkmark \\
        \hline
        TAG~\cite{Google2021MTLG} & \checkmark & - & - \\
        \hline
        PCGrad~\cite{kurin2022drawbcPCGRAD} & \checkmark & - & - \\
        \hline
        Recon~\cite{guangyuan2022recon} & \checkmark & - & \checkmark \\
        \hline
        L2TFMT~\cite{RoyNtoutsiECML22} & - & \checkmark & - \\
        \hline
        WB-fair~\cite{hu2023fairMTL} & - & \checkmark & - \\
        \hline
        \our{} & \checkmark & \checkmark & \checkmark\\
        \hline
    \end{tabular}
\end{table}
 \section{Background And Motivation}\label{sec.basics}
\subsection{Background Setup}\label{sec.basics}
We assume a dataset $D=X\times S \times Y$
consisting of $m$-dimensional \emph{non-protected attributes} $X\in \mathbb{R}^{m}$, \emph{protected attribute} $S\in \mathbb{S}$, and an output part $Y=Y_1 \times \cdots \times Y_T$ referring to the associated class labels for the output tasks $1, \cdots, T$. 
For simplicity, we assume binary tasks, i.e., $Y_t \in \{0,1\}$, $t=1,\cdots,T$; with $1$ representing a positive (e.g., ``granted") and $0$ representing a negative (e.g.,``rejected") class, and a binary protected attribute: $\mathbb{S}= \{g,\overline{g}\}$, where $g$ and $\overline{g}$ represent demographic groups like ``female", and ``male".

Let $\mathcal{M}$ be a deep \verb|MTL| model with ($d$+1) layers, 
parameterized by the set $\theta \in \Theta$ of parameters, which includes: $d$ layers of shared parameters $\theta_{sh}$ (i.e., weights of layers shared by all tasks) connected in order of depth from 1 to $d$, and for every task $t$ a single layer of task-specific parameters $\theta_t^{d+1}$ (i.e., weights of the task specific layers) connected to the topmost shared layer $\theta_{sh}^d$. Formally, we describe the parameters of $\mathcal{M}$ as  $\theta=\theta_{sh}^{1,\cdots,d}\times \theta_1^{d+1}\times\cdots\times\theta_T^{d+1}$, where $\theta_{sh}^{1,\cdots,d}$ indicates that shared parameters extends from depth $1$ to $d$, we use $\theta_{\alpha}^{b}$ to indicate any parameters $\theta_{\alpha}$ at a certain depth $b$.   

Typically, in fair-\verb|MTL| for every task $t$, $t=1,\cdots,T$, the goal is to minimize an accuracy loss function $\mathcal{L}_t()$, and a fairness loss function $\mathcal{F}_t()$.  In this work, for $\mathcal{L}_t()$ we use the  negative log likelihood, and for $\mathcal{F}_t()$ the robust log~\cite{rezaei2020robustlogloss,RoyNtoutsiECML22}: 
\small\begin{equation}\label{eq:task_losses}
\begin{split}
    \mathcal{L}_t(\theta,X)=&-Y_t \log \mathcal{M}^t(X,\theta) - 
    (1-Y_t) \log (1-\mathcal{M}^t(X,\theta)))
  \\
  \mathcal{F}_t(\theta,X,S)&= \sum_{y\in\{1,0\}}\max\big( \mathcal{L}_t(\theta,X\mid Y_t=y,S=g), \\
  & \phantom{1234577}\mathcal{L}_t(\theta,X\mid Y_t=y,S=\overline{g})\big)
\end{split}
\end{equation}\normalsize
 where $\mathcal{M}^t(X,\theta)$ is the outcome $\mathcal{M}$ on task $t$ based on model parameters $\theta$. Note that $\mathcal{F}$ uses an operator ($max$) over several $\mathcal{L}$ conditioned on different demographics ($g$, $\overline{g}$), that enables $\mathcal{M}$ to emphasise the demographic group on which it makes the maximum likelihood error. Further, we denote 
$\nabla_{\theta_\alpha}^{\mathcal{L}_t}$, and $\nabla_{\theta_\alpha}^{\mathcal{F}_t}$ as the gradient for parameter $\theta_\alpha$ w.r.t., loss $\mathcal{L}_t$, and $\mathcal{F}_t$ resp. on task $t$. 

The unfairness of $\mathcal{M}$ on task $t$ can be measured based on the generic framework by~\cite{roy_facct2023}
as the absolute difference in predictions between $g$ and $\overline{g}$ under a given set of conditions $\mathbb{C}$ which govern the type of fairness definition used:
\small\begin{equation}\label{eq.fair_gen}
 F_{viol}^{(t)}(\mathcal{M}) =  \sum_{c\in \mathbb{C}}|P(\mathcal{M}^t(X)|S=g,c)-P(\mathcal{M}^t(X)|S=\overline{g},c)|
\end{equation}\normalsize 
In this work, we adopt two popular fairness measures~\cite{hardt2016equality}:\\ \small{\em Equal Opportunity} ($ EP_{viol}^{(t)}(\mathcal{M})$) where $\mathbb{C}: \{[\mathcal{M}^t(X)=1|Y_t=1]\}$, and {\em Equalized Odds} ($EO_{viol}^{(t)}(\mathcal{M})$) where $\mathbb{C}: \{[\mathcal{M}^t(X)=1,Y_t=1],[\mathcal{M}^t(X)=0,Y_t=1]\}$. \normalsize
EP ephasizes fairness in the positive class, 
while EO considers fairness across all classes.

\subsection{Negative Transfer and Gradient Conflict}\label{sec.conflicts}
The term \nt{} is akin to the concept of negative knowledge gain. Knowledge gain ($KG$) on a task $t$ by any MTL model $\mathcal{M}$ is assessed as the difference in accuracy between $\mathcal{M}$ and a single-task learner (STL) $\mathcal{H}$ trained on $t$:
\small\begin{equation}\label{eq.kt}
KG(t): P(\mathcal{M}^t(X)=Y_t)-P(\mathcal{H}(X)=Y_t)
\end{equation}\normalsize
The ideal scenario is to achieve a positive (or at least non-negative) transfer, i.e., $KG(t)\geq0$ for all tasks. Any failure to meet this condition is termed as \nt{}, where $KG(t)<0$. 
Research into conflicting gradients has identified accuracy conflict as the root cause of the \nt{} problem \cite{du2018gradient_similarity,guangyuan2022recon,yu2020surgery}. Accuracy conflict between any two task gradients $\nabla_{\theta_\alpha}^{\mathcal{L}{t_1}}$ and $\nabla{\theta_\alpha}^{\mathcal{L}{t_2}}$ is defined as:
\small\begin{equation}\label{eq.task_conflicts}
\begin{aligned}
&conflict(\nabla{\theta_\alpha}^{\mathcal{L}{t_1}},\nabla{\theta_\alpha}^{\mathcal{L}{t_2}}):
\nabla{\theta_\alpha}^{\mathcal{L}{t_1}}\boldsymbol{\cdot}\nabla{\theta_\alpha}^{\mathcal{L}{t_2}}<0
\end{aligned}
\end{equation}\normalsize
It follows from Eq.~\ref{eq.task_conflicts} that accuracy conflict happens when \small${\frac{\pi}{2}<\measuredangle{(\nabla_{\theta_\alpha}^{\mathcal{L}_{t_1}},\nabla_{\theta_\alpha}^{\mathcal{L}_{t_2}})}<-\frac{\pi}{2}}$\normalsize.

\subsection{Bias Transfer and Fairness Conflict} 
Following the idea of knowledge gain (Eq.~\ref{eq.kt}), we define the concept of discrimination gain ($DG$) for a given task $t$ as the difference of fairness violation for any \verb|MTL| $\mathcal{M}$ against an \verb|STL| $\mathcal{H}$ on task $t$: 
\small\begin{equation}\label{eq.unfair_transfer}
  DG(t): F_{viol}^{(t)}(\mathcal{M})- F_{viol}^{(t)}(\mathcal{H})
\end{equation}\normalsize 
We say a negative gain of fairness \emph{aka} \ut{} is observed when $DG(t)>0$ for any given $t$. 
Notice that contrary to \nt{}, the condition for \ut{} is attained when the left part of Eq~\ref{eq.unfair_transfer} is positive. This is because ideally we want the bias of the \verb|MTL| to be lower than that of \verb|STL|. 

We hypothesize that similar to \nt{}, \ut{} is induced by a gradient conflict, which we term as \textbf{fairness conflict},  \small $conflict(\nabla_{\theta_\alpha}^{\mathcal{F}_{t_1}},\nabla_{\theta_\alpha}^{\mathcal{F}_{t_2}}): \nabla_{\theta_\alpha}^{\mathcal{F}_{t_1}}\boldsymbol{\cdot}\nabla_{\theta_\alpha}^{\mathcal{F}_{t_2}}<0$\normalsize. 

Our aim is to ensure \nt{} free and \ut{} free learning of an \verb|MTL| $\mathcal{M}$ by ensuring conflict free learning for both accuracy and fairness. Now, unrolling the gradient update to a parameter $\theta_\alpha$ for losses $\mathcal{L}$ and $\mathcal{F}$ of any two tasks $t_1$ and $t_2$, learned by a vanilla fair-\verb|MTL| with a learning rate $\eta$, we have:
\small\begin{equation}\label{eq.grad_unroll}
\begin{split}
    \theta_{\alpha} \leftarrow \theta_{\alpha} - \eta &\sum_{t\in\{t_1,t_2\}} \nabla_{\theta_{\alpha}}(\mathcal{L}_t + \lambda_t \mathcal{F}_t)
    = \theta_{\alpha} -\eta(\nabla_{\theta_\alpha}^{\mathcal{L}_{t_1}} \\& + \nabla_{\theta_\alpha}^{\mathcal{L}_{t_2}}) 
    - \eta(\lambda_{t_1}\nabla_{\theta_\alpha}^{\mathcal{F}_{t_1}} + \lambda_{t_2}\nabla_{\theta_\alpha}^{\mathcal{F}_{t_2}})
\end{split}
\end{equation}\normalsize
Now, from Eq.~\ref{eq.grad_unroll} we infer that for any two tasks we can tackle the accuracy conflict and fairness conflict  separately.

\section{FairBranch}
\label{sec.algo}

Our method, \our{}, addresses both negative transfer (\nt{}) and unfair transfer (\ut{}). In Algorithm~\ref{algo:fairbranch}, we initialize each task $t$ as a task-group $\{t\}$, with task-specific parameters $\theta_t$ and shared parameters $\theta_{sh}^{1,\cdots,d}$ of $d$ layers. $TG$ denotes the collection of all such task-groups. At each training loop, we compute task gradients $\nabla_{\theta_\alpha}^{\mathcal{L}{t}}$ and $\lambda_t\nabla{\theta_\alpha}^{\mathcal{F}{t}}$ for each task $t=1,\cdots,T$ (line 2). Here, $\lambda_t$ is an intra-task weight addressing accuracy-fairness conflicts, set to 0 when \small$ \nabla{\theta_{t}}^{\mathcal{L}{t}}\boldsymbol{\cdot}\nabla{\theta_{t}}^{\mathcal{F}_{t}}<0$\normalsize.

We then correct fairness conflicts (FBGrad) in each task branch (line 3). After updating $\mathcal{M}$'s parameters (line 4), we check the $d_c$ layer for conditions (line 5) and cluster similar task-groups within $TG$ based on branch parameter similarities at $d_c-1$ (line 6). We merge task-groups in each cluster by forming branch parameters at $d_c$ exclusive to the cluster, minimizing accuracy conflicts and \nt{}. Finally, we update $d_c$ (line 7) and continue training. We detail the Branching mechanism 
at current depth $d_c$ (Sec.~\ref{sec.branching}), and Fairness-conflict correction mechanism on each branch parameter in $\mathcal{M}$ (Sec.\ref{sec.fbgrad}). 

\begin{algorithm}[!htbp]\caption{The FairBranch algorithm}\label{algo:fairbranch}
\small
	\algorithmicrequire{$D=\{(x_i,s_i,y_i^1, \cdots y_i^T )\}_{i=1}^n$, $\mathcal{M}$  parameters: $\theta=\theta_{sh}^{1,\cdots,d}\times \theta_1^{d+1}\times\cdots\times\theta_T^{d+1}$ 
 }
	\\ \textit{Initialisation} :  current layer depth: $d_c\leftarrow d$, $e\leftarrow 0$, ${TG}\leftarrow \{\{1\},\cdots,\{T\}\}$ 
	\begin{algorithmic}[1]
 \STATE
     \textbf{Until} $\{\mathcal{L}_t\}$ and $ \{\mathcal{F}_t\}$ convergence do $e\leftarrow e+1$ 
	    \STATE  compute task gradients $\nabla_{\theta_\alpha}^{\mathcal{L}_{t}} $, $\lambda_t\nabla_{\theta_\alpha}^{\mathcal{F}_{t}}$;   $\forall\theta_\alpha \in \theta$, $t=1,\cdots T$
        \STATE  FBGrad: apply fairness-conflict correction on branches.
        \STATE  Update all $\theta_\alpha\in \theta$ (c.f., Eq~\ref{eq.grad_unroll})
        \STATE  \textbf{if}  \text{branch condition} is True
        \STATE  apply branching mechanism on $\mathcal{M}$
        \STATE  $d_c\leftarrow d_c-1$
        \STATE \textbf{End if}
        
        \STATE \textbf{End Until}
	\end{algorithmic}
	\algorithmicensure{fair-Branch MTL $\mathcal{M}$} 
\end{algorithm}

\subsection{Branching Mechanism}\label{sec.branching}
\begin{figure}
    \centering
    \includegraphics[width=0.9\linewidth]{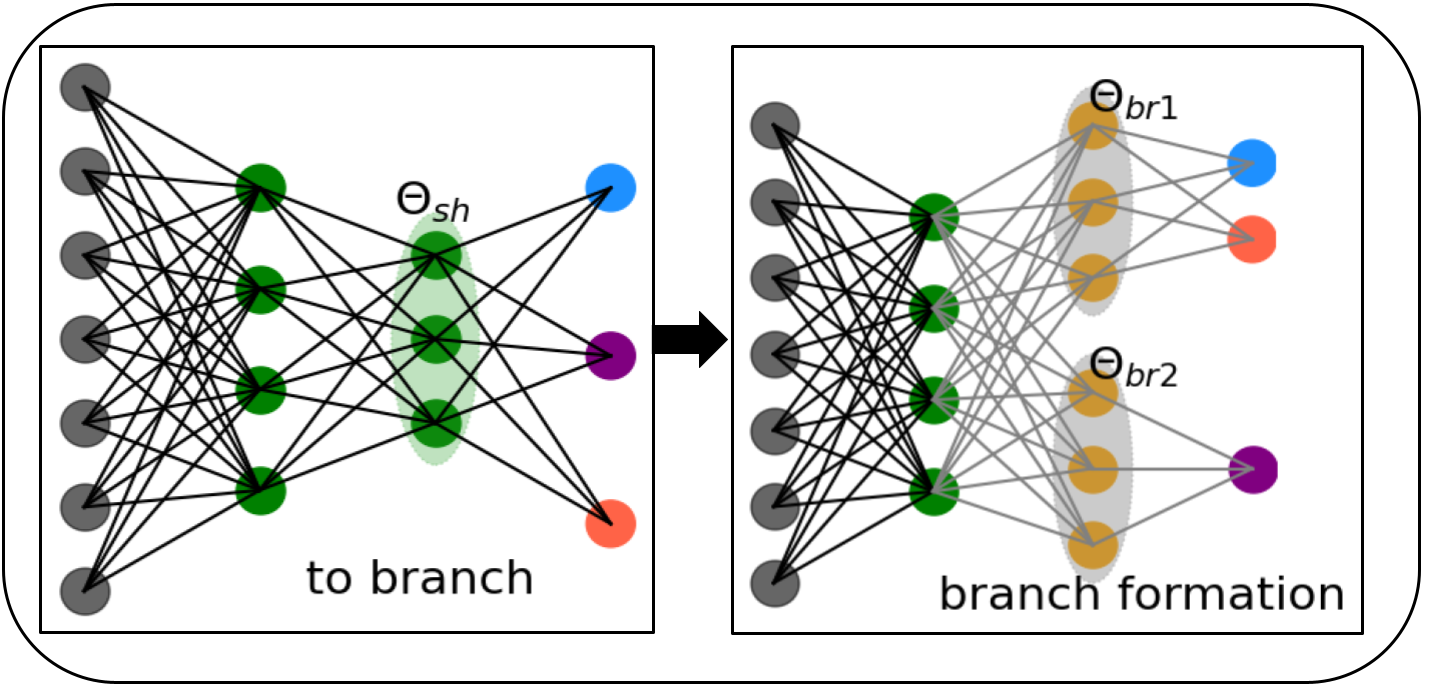}
    \vspace{-0.05in}
    \caption{\small A High Level Depiction of Branch Formation}
    \label{fig:fairbranch}
\end{figure}


In \our{}, branching occurs only if two conditions are met: a) $|{TG}|\geq 2$, indicating multiple groups within $TG$, and b) $d_c > 1$, meaning the current depth is not at the input layer. Once both conditions hold true, branching 

\subsubsection{Measuring task-group affinity:}
The pairwise affinity between the task-groups within $TG$ is measured using a parameter similarity function $sim()$ and a threshold hyperparameter $\tau\in (0,1]$.  
\small\begin{equation}\label{eq.affinity}
\mathcal{A}\leftarrow \{ v|v= sim(\theta_{\alpha}^{d_c+1},\theta_{\beta}^{d_c+1})\geq \tau; \alpha,\beta \in {TG}; \alpha\neq \beta \}
\end{equation}\normalsize
The idea is to cluster together only task-groups pairs that have similarity higher than or equal to the given threshold. The similarity function ($sim()$) that we use in \our{} is based on central kernel alignment~\cite{cortes2012algorithms}, which has gained recent popularity in parameter similarity measures~\cite{Hinton2019similarity,tang2020similarity,csiszarik2021similarity} due to its desired invariant properties\cite{Hinton2019similarity}. Formally, it is defined as:
\small\begin{equation}\label{eq.similarity_func}
      sim(\theta_{\alpha},\theta_{\beta})= cka(K(\theta_{\alpha}),K(\theta_{\beta}))
\end{equation}\normalsize
where $\theta_{\alpha},\theta_{\beta}$ are branch parameters exclusive to task-groups $\alpha$ and $\beta$ respectively at the layer depth just one above the current depth $d_c$, $K(\theta_\alpha)= \theta_{\alpha}\theta_{\alpha}^\intercal$ is a linear kernel function with $\theta_\alpha^\intercal$ as the transpose of $\theta_\alpha$, and $cka$ is kernel alignment measure defined as:
\small\begin{equation}\label{eq.cka}
      cka(K_\alpha,K_\beta)=
         \frac{tr(\mathcal{I}(K_\alpha) \mathcal{I}(K_\beta))}{\sqrt{tr(\mathcal{I}(K_\alpha) \mathcal{I}(K_\alpha))tr(\mathcal{I}(K_\beta) \mathcal{I}(K_\beta))}}
\end{equation}\normalsize
where $\mathcal{I}$ is a centering function~\cite{o2021centering}, $K_\alpha$ is $K(\theta_\alpha)$, and $tr()$ denotes trace of the resultant centred matrix. 
\subsubsection{Clustering on affinity}
We use the affinities $\mathcal{A}$ (Eq.~\ref{eq.affinity}) to cluster the task-groups in $TG$. 
Although our algorithm offers flexibility in the selection of the clustering method, in our implementation we opted for the Single Linkage Hierarchical Clustering (SLHC)~\cite{xu2005clustsurvey}. We start with an empty cluster $\mathcal{C}=\emptyset$, and then recursively include task (or task group) pairs in $\mathcal{C}$, greedily on the basis of $\mathcal{A}$ until $\mathcal{A}$ is $\emptyset$:
\small\begin{equation}\label{eq.clust}
\begin{split}  
 \text{Until } \mathcal{A}\neq \emptyset: \mathcal{C}\leftarrow \mathcal{C} \cup \{\{\Tilde{\alpha},\Tilde{\beta}\}\}|\Tilde{\alpha},\Tilde{\beta}=\argmax\limits_{\alpha,\beta\in {TG}} \mathcal{A}; \\ 
  \mathcal{A}\leftarrow\mathcal{A}/\{\{sim(\alpha,\beta)|\alpha=\Tilde{\alpha} \lor \beta=\Tilde{\beta}; \alpha,\beta \in {TG} \}
\end{split}
\end{equation}\normalsize
The tasks (or tasks-groups)  that are not included in $\mathcal{C}$ are added in $TG$ as a singleton. 
 The main motivation behind finding such binary task groups is to limit the scope of the number of possible conflicts (both accuracy and fairness) between task groups in any given branch, 
which enables the model to efficiently scale to a large number of tasks. 
\small\begin{equation}\label{eq.tg_update}
    TG\leftarrow Cluster(TG,\mathcal{A}) \cup \{\{\gamma\}|sim(\theta_\gamma,\theta_\alpha) \notin \mathcal{C}, \gamma,\alpha \in {TG}\}
\end{equation}\normalsize

\subsubsection{Branch formation}
Next, we use the updated task-groups $TG$ (Eq.~\ref{eq.tg_update}) to form the branches ${br}^{d_c}$. Branches are a collection of parameters \small${br}^{d_c}=\{\theta_{br_t}|\theta_{br_t}=copy(\theta_{sh}^{d_c});t=1,\cdots,|TG|\}$\normalsize, where every parameter $\theta_{br_t}$ initiates with a replica of the shared parameter $\theta_{sh}^{d_c}$ which is currently being branched. In $\mathcal{M}$, we replace the parameter $\theta_{sh}^{d_c}$ with the parameters collection ${br}^{d_c}$, and connect each $\theta_{br_t}\in {br}^{d_c}$ with $\theta_{sh}^{d-1}$ below. Based on the above, each $\theta_{br_t}$ is connected with a unique parameter pair $\theta_\alpha^{d_c+1}$, $\theta_\alpha^{d_c+1}$ s.t.,  $(\alpha,\beta)\in \Tilde{p}$, \small $\exists!\Tilde{p}\in TG$\normalsize. Fig.~\ref{algo:fairbranch} depicts a toy example to highlight the architectural change $\mathcal{M}$ undergoes by forming new branches. 

\subsection{Fairness Conflict Correction}\label{sec.fbgrad}
Denoted by FBGrad, in our FairBranch algorithm this step is responsible to mitigate \ut{}. This step is highly motivated from PCGrad update \cite{kurin2022drawbcPCGRAD} for correcting gradient conflicts. The key difference here is that for the fairness gradient correction instead of the adjusting the gradients at every layer, we look only into the layers that have been branched. The intuition is to apply fairness correction only on parameters without any \nt{}, to limit the scope of cross-task fairness-accuracy conflicts (this problem is discussed in detail in Sec.~\ref{sec.theory}). 
To execute this step we look into each of the branch parameters $\theta_{br_t}^b\in br^{b}$ and $b$ runs from $d$ to $,d_c$, and identify the tasks $(t_1,t_2,\cdots)$ connected to $\theta_{br_t}^b$.

For each pair (if any) of tasks $t_1$ and $t_2$ connected with any branch $\theta_{br}$, we check for fairness conflicts between $\nabla_{\theta_{br}}^{\mathcal{F}_{t_1}}$ and $\nabla_{\theta_{br}}^{\mathcal{F}_{t_2}}$ (c.f. Sec~\ref{sec.conflicts}). {\em Iff} conflict is found we correct both the task gradients by FBGrad function w.r.t. one another, where update of $\nabla_{\theta_{br}}^{\mathcal{F}_{t_1}}$ w.r.t $\nabla_{\theta_{br}}^{\mathcal{F}_{t_2}}$ is defined as:
\small\begin{equation}\label{eq.grad_rejection}
  FBGrad:  \nabla_{\theta_{br}}^{\mathcal{F}_{t_1}}= \nabla_{\theta_{br}}^{\mathcal{F}_{t_1}} - \frac{{\nabla_{\theta_{br}}^{\mathcal{F}_{t_1}}}\boldsymbol{\cdot}\nabla_{\theta_{br}}^{\mathcal{F}_{t_2}}}{\|\nabla_{\theta_{br}}^{\mathcal{F}_{t_2}}\|^2}\nabla_{\theta_{br}}^{\mathcal{F}_{t_2}}
\end{equation}\normalsize

\section{Theoretical Analysis}\label{sec.theory}

\subsection{Why Parameter Similarity?}
Let us assume any two tasks $t_1$ and $t_2$, are identified to form a group by FairBranch. Now, using the Hilbert-Schmidt Independence criterion~\cite{gretton2005hsic} and Eq.~\ref{eq.affinity} and \ref{eq.similarity_func}, we get
\begin{equation}\label{eq.hsic}
\small
    sim(\theta_{t_1},\theta_{t_2})\geq \tau \implies \frac{{\|\theta_{t_2}^{\intercal}\theta_{t_1}\|}_{\mathbb{F}}^2}{{\|\theta_{t_1}^{\intercal}\theta_{t_1}\|}_{\mathbb{F}}{\|\theta_{t_2}^{\intercal}\theta_{t_2}\|}_{\mathbb{F}}}\geq \tau
\end{equation}
where ${\|\cdot\|}_{\mathbb{F}}$ is the Hilbert-Schmidt norm.  Since, we choose $\tau>0$, 
we have $tr(\theta_{t_2}^\intercal\theta_{t_1})>0$ i.e. $\theta_{t_1}\boldsymbol{\cdot}\theta_{t_2}>0$. 
Thus, $\theta_{t_1}$ and $\theta_{t_2}$ are moving in  similar directions. 

Let $\mathcal{L}_t$ be a Lipschitz continuous and convex~\cite{boyd2004convex} task loss, and $\theta_{t}^{(0)}=C$ be the initial parameters for task $t$. Also let $\mathcal{L}_t(j)$ be the loss, and $\theta_t{(j)}$ be the parameter for $t$ at $j$-th epoch; then after $e$ epochs we have: 
\begin{equation}\label{eq.grad_update}
\small
    \begin{split}
        \theta_{t}{(e)} \leftarrow \theta_{t}{(e-1)} - \eta \nabla_{\theta_{t}{(e-1)}} ^{\mathcal{L}_{t}(e)}
        =\theta_t{(0)} - \eta\sum_{j=1}^{e}\nabla_{\theta_t{(j-1)}}^{\mathcal{L}_t(j)}
    \end{split}
\end{equation}
Since $\theta_{t_1}{(0)}=\theta_{t_2}{(0)}=\mathcal{C}$, geometrically we assume $\theta_{t_1}{(0)}$ and $\theta_{t_2}{(0)}$ to be the common starting point of $t_1$ and $t_2$ (say (0,0) in 2D co-ordinates). Without loss of generality, we can say that $\sum_{j=1}^{e}\nabla_{\theta_t{(j-1)}}\mathcal{L}_t(j)$ is the resulting gradient vector of all the gradients observed till epoch $e$ for task $t$. Thus, we can infer that $\sum_{j=1}^{e}\nabla_{\theta_{t_1}{(j-1)}}\mathcal{L}_{t_1}(j)\boldsymbol{\cdot}\sum_{j=1}^{e}\nabla_{\theta_{t_2}{(j-1)}}\mathcal{L}_{t_2}(j)>0$ when $sim(\theta_{t_1},\theta_{t_2})\geq \tau$, i.e. the resulting gradient movement for both tasks is in a similar direction. 
Since, such resulting gradients are accumulated over multiple batches of the data, it is expected to be stable and give a strong estimation of the direction of minima.  Henceforth, our intuition is that given a strong similarity ($\tau\rightarrow 1$), we can ensure that the direction of minima of two tasks $t_1$ and $t_2$ is similar when $sim(\theta_{t_1},\theta_{t_2})\geq \tau$, and thus is expected to move together without any conflict.  

\subsection{Why Only Branch Specific Fairness Correction?}

\begin{figure}
    \centering
    \begin{subfigure}[b]{0.49\linewidth}
    \centering
\includegraphics[width=0.8\columnwidth]{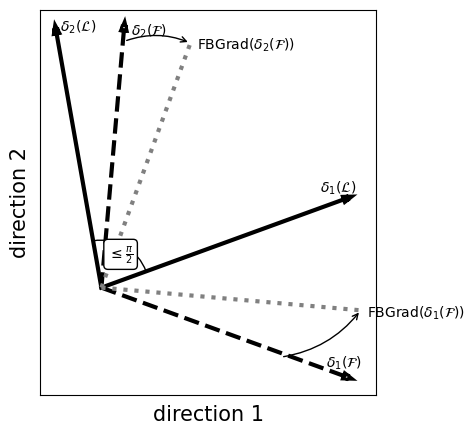}
    \end{subfigure}
    \begin{subfigure}[b]{0.49\linewidth}
    \centering
\includegraphics[width=0.9\columnwidth]{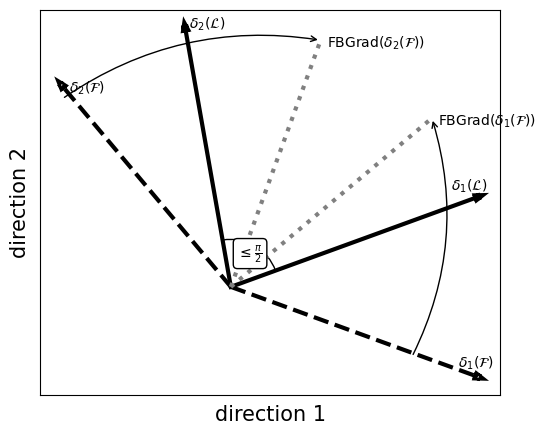}
    \end{subfigure}
    \caption{Example Showing Effect of Fairness Gradient Correction on Task-grouped Branches.}
    \label{fig:exmpl_correction}
\end{figure}

In fair-MTL frameworks, at least two different losses($\mathcal{L}_t$ and $\mathcal{F}t$) per task $t$ are accommodated, 
which can lead to conflicts when gradients from these losses disagree with each other in the direction of update. With $T$ tasks, a fair-\verb|MTL| has the potential for $T(2T-1)$ conflicts at every $\theta_{sh}^d$ layer. 
For instance, with only two tasks $t_1$ and $t_2$, there are $T(2T-1)=6$ conflicts, including four inter-task conflicts ($(\nabla_{\mathcal{L}_1},\nabla_{\mathcal{L}_2)}$, $(\nabla_{\mathcal{F}_1},\nabla_{\mathcal{F}_2})$, $(\nabla_{\mathcal{L}_1},\nabla_{\mathcal{F}_2})$, $(\nabla_{\mathcal{L}_2},\nabla_{\mathcal{F}_1})$), and two intra-task conflicts ($( \nabla_{\mathcal{L}_1},\nabla_{\mathcal{L}_2})$). 
In our algorithm (Sec.~\ref{sec.algo}), we address intra-task conflicts by imposing a strong condition on $\lambda_t$ and handle inter-task accuracy conflicts ($( \nabla_{\mathcal{L}_1},\nabla_{\mathcal{L}_2})$) in the created branches by grouping related tasks. Thus, in the branched layers, applying FBGrad not only resolves fairness conflicts ($( \nabla_{\mathcal{F}_1},\nabla_{\mathcal{F}_2})$) but also reduces the likelihood of inter-task fairness-accuracy conflicts ($( \nabla_{\mathcal{L}_1},\nabla_{\mathcal{F}_2})$, $(\nabla_{\mathcal{L}_2},\nabla_{\mathcal{F}_1})$) in most scenarios. 
Examining potential post-conflict correction scenarios within branches unveils five possibilities, with four leading to FBGrad projecting fairness gradients towards zones free of inter-task fairness-accuracy conflicts. Two illustrative hypothetical examples, showcased in Fig~\ref{fig:exmpl_correction}, demonstrate these scenarios.
However, in shared layers, lacking a branch mechanism precludes such assurances. Correcting fairness conflicts in shared layers might inadvertently worsen inter-task fairness-accuracy conflicts. Our experimental findings will showcase how fairness conflict correction in task-group branches effectively mitigates negative and biased transfer.

\color{black}
\section{Experiments}

\noindent \textbf{Datasets}: 
We conduct experiments on two datasets across four setups. The first two setups use tabular data from the \textit{ACS-PUMS} dataset~\cite{ding2021retiring}, following a protocol of training on one year and testing on the next~\cite{RoyNtoutsiECML22}. The setups are: i) \textit{ACS-PUMS 18-19}, trained on 2018 and tested on 2019 census data, and ii) \textit{ACS-PUMS 19-21}, trained on 2019 and tested on the latest available 2021 census data. We use gender as the protected attribute in both setups.
The next two setups are based on the CelebA dataset~\cite{liu2015celebA}, consisting of celebrity face images. We follow the provided training-test partition. Adopting an existing fair-MTL protocol~\cite{RoyNtoutsiECML22}, we create two experiment setups: i) \textit{CelebA gen} with 17 tasks and gender as the protected attribute, and ii) \textit{CelebA age} with 31 tasks and age as the protected attribute.
\noindent \textbf{Competitors:} We compare \our{} with six state-of-the-art MTL methods. The MTL competitors are selected from every direction that our work covers:

\begin{itemize}
   
 \item Task-grouping: i) \textbf{FAFS}~\cite{lu2017FAFS}, and ii) \textbf{TAG}~\cite{Google2021MTLG}.

 \item Conflict-aware: iii) \textbf{PCGrad}~\cite{yu2020surgery}, and iv) \textbf{Recon}~\cite{guangyuan2022recon}.

 \item Fairness-aware:  v) \textbf{L2TFMT}~\cite{RoyNtoutsiECML22} and vi) \textbf{WB-fair}~\cite{hu2023fairMTL}. 
\end{itemize}
We implement the methods in their vanilla form\footnote{No fairness correction for FAFS, TAG, PCGrad, and Recon.}. 

\noindent \textbf{Evaluation Measures}\label{sec:evalMeasures}: For comparative overall evaluation in Sec.~\ref{sec.results}, we report the average knowledge gain $\bar{KG}=\frac{1}{T}\sum_t KG(t)$ (Eq.~\ref{eq.kt}) for \nt{}, and average discrimination gain $\bar{DG}=\frac{1}{T}\sum_t DG(t)$ (Eq.~\ref{eq.unfair_transfer}) for \ut{}.  
Then, to obtain in-depth per-task performance comparison of the models, we plot the negative and bias transfer distribution over the tasks. 
For qualitative analysis of \our{} in tackling  negative and bias transfer, in Sec.~\ref{sec.result_conflict} we present the distribution of fairness conflicts and accuracy conflicts of the learned gradients observed between the tasks while training. To understand which tasks have the most conflicts over the training, we plot the cross-task conflict heat-maps. 

\noindent \textbf{Hyperparameters}: For tabular setups, we use $\tau=0.7$, we split the training data into 70:30 training:validation, stratified across all census states. For computer vision setups, we use $\tau=0.8$, and the predefined training:validation:test split~\cite{liu2015celebA}.
\begin{figure*}[t]
    \centering
     \begin{subfigure}[b]{0.49\linewidth}
         \centering
         \includegraphics[width=\columnwidth]{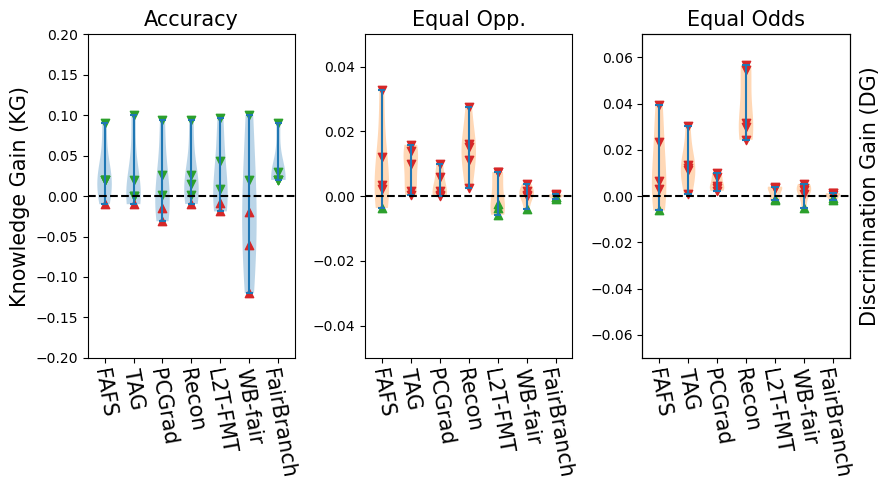}
         \vspace{-.3in}
         \caption{ACS-PUMS 18-19}\label{fig.pums1}
    \end{subfigure}
    \begin{subfigure}[b]{0.49\linewidth}
         \centering
         \includegraphics[width=\columnwidth]{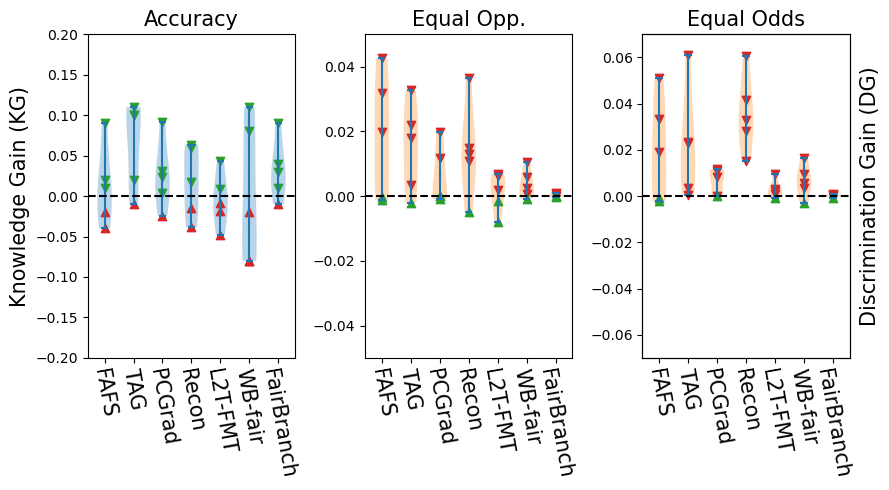}
         \vspace{-.3in}
         \caption{ACS-PUMS 19-21}\label{fig.pums2}
    \end{subfigure}
    \begin{subfigure}[b]{0.49\linewidth}
         \centering
         \includegraphics[width=\columnwidth]{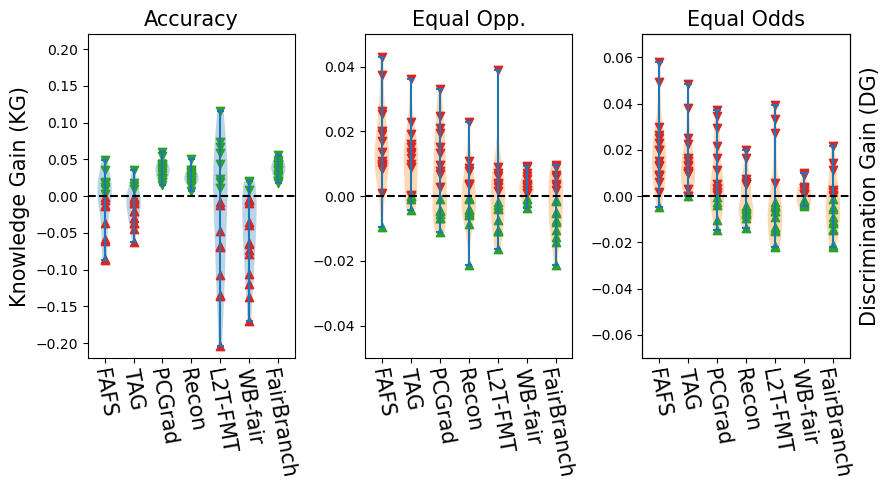}
         \vspace{-.3in}
         \caption{CelebA gen}\label{fig.celeb1}
    \end{subfigure}
    \begin{subfigure}[b]{0.49\linewidth}
         \centering
         \includegraphics[width=\columnwidth]{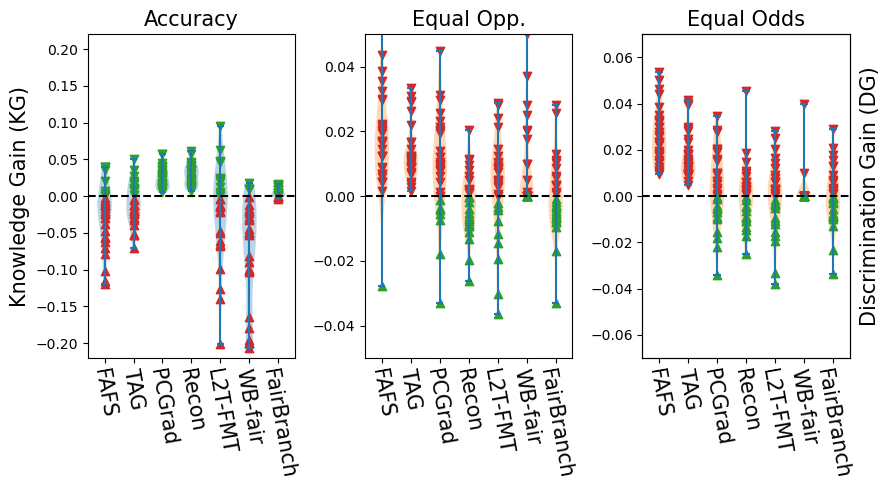}
         \vspace{-.3in}
         \caption{CelebA age}\label{fig.celeb2}
        
    \end{subfigure}
    \caption{\small Comparison on Knowledge Gain (KG) and Discrimination Gain (DG) Distribution: Each box provides comparison on a given Metric Labelled on Top. In boxes every triangle depicts Difference between an MTL with Task Specific STLs. Red Triangles indicates Negative/Bias Transfer and Green indicates Positive/Unbiased Gain. Positive Difference for Accuracy, Negative for Fairness are better.}
    \label{fig:overall}
\end{figure*}
\begin{figure*}
    \centering
     \begin{subfigure}[b]{\linewidth}
         \centering
         \includegraphics[width=0.9\columnwidth]{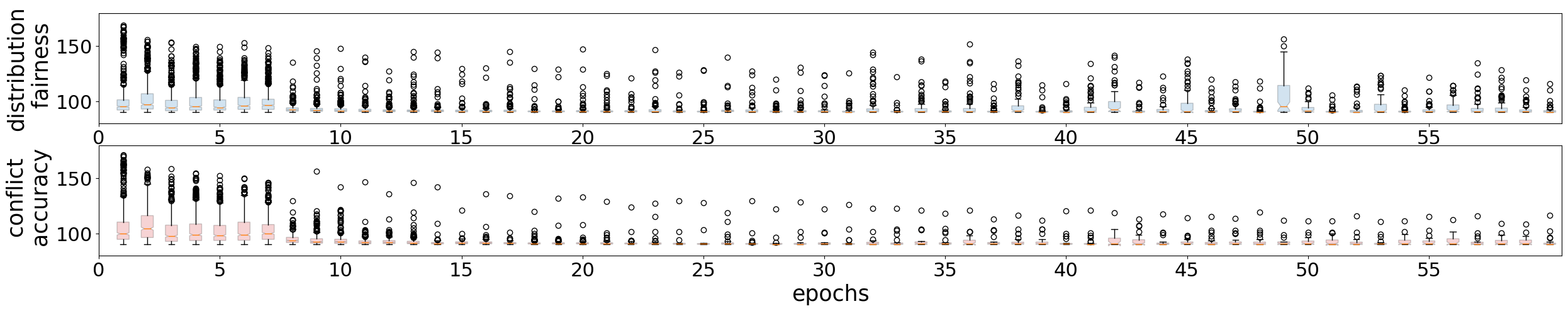}
         \vspace{-.1in}
         \caption{CelebA gen}\label{fig.dist_gen}
    \end{subfigure}
    \begin{subfigure}[b]{\linewidth}
         \centering
         \includegraphics[width=0.9\columnwidth]{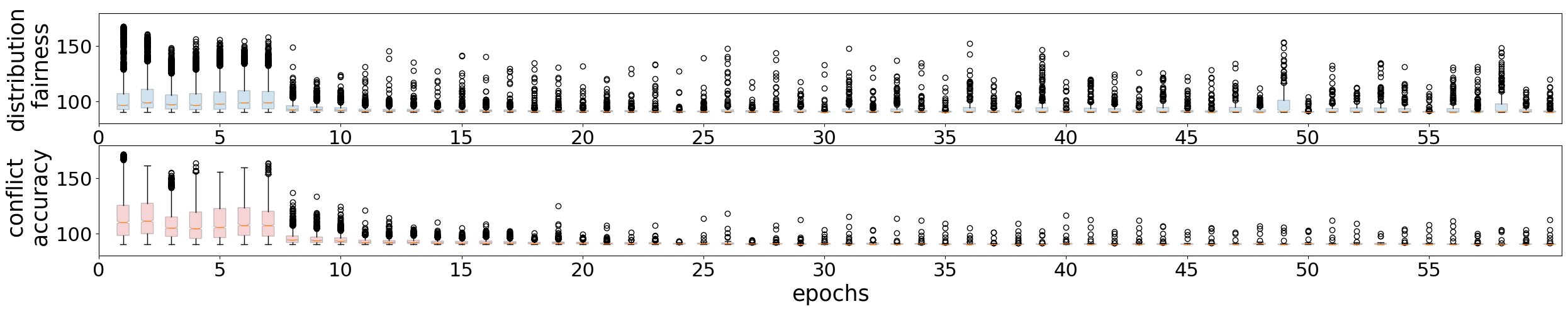}
         \vspace{-.1in}
         \caption{CelebA age}\label{fig.dist_age}
    \end{subfigure}
    \caption{\small Accuracy and Fairness Loss Gradient Conflicts of FairBranch  over Training Epochs. Each Box shows Distribution of Angle of Conflict Observed at an Epoch. Less Densely Crowded Lower Boxes are Better.}\label{fig.conflicts_dist}
    \vspace{-0.1in}
\end{figure*}
\subsection{Comparative Results}\label{sec.results}

\begin{table}
\centering
\small 
\caption{\small Comparative Results: $\bar{KG}$ (Higher is better) for Accuracy (Negative Values indicates Negative Transfer), and $\bar{DG}$ (Lower is better) for Fairness (Positive Values indicates Bias Transfer). Best Values in Gray Cell, Second Best underlined.}
\label{tab:exp}
\setlength{\tabcolsep}{5pt} 
\renewcommand{\arraystretch}{1.25}
\begin{tabular}{|rc|l|c c|c c|}

\hline
&\textbf{Model} & \textbf{Metric} & \multicolumn{2}{c|}{\textbf{ACS-PUMS}} & \multicolumn{2}{c|}{\textbf{CelebA}}\\
&& &\textbf{18-19} & \textbf{19-21} & \textbf{gen} & \textbf{age} \\
\hline
\multirow{6}{*}{\rotatebox{90}{\footnotesize\textbf{Task-grouping}}} &\multirow{3}{*}{FAFS} & $\bar{\textit{KG}}$ & \underline{0.028} & 0.012 & -0.011 &-0.024\\
& & \multirow{2}{*}{$\bar{\textit{DG}}$ \begin{tabular}[c]{@{}|c@{}}EP\\EO\end{tabular}} & 0.009& 0.019 & 0.015 & 0.017\\
& &  & 0.013 & 0.020 & 0.019 & 0.026\\
\cline{3-7}
&\multirow{3}{*}{TAG} & $\bar{\textit{KG}}$ & 0.022 & \cellcolor{gray!35}{\textbf{0.064}} & -0.012 & -0.010\\
&& \multirow{2}{*}{$\bar{\textit{DG}}$ \begin{tabular}[c]{@{}|c@{}}EP\\EO\end{tabular}} & 0.008 & 0.015 & 0.015 & 0.013\\
&&  & 0.014 & 0.022 & 0.010 & 0.017\\
\hline
\multirow{6}{*}{\rotatebox{90}{\footnotesize\textbf{Conflict aware}}}&\multirow{3}{*}{PCGrad} & $\bar{\textit{KG}}$ & 0.015& 0.025& \underline{0.035} &  \underline{0.025}\\
&& \multirow{2}{*}{$\bar{\textit{DG}}$ \begin{tabular}[c]{@{}|c@{}}EP\\EO\end{tabular}} &0.004 & 0.006& 0.007 & 0.009\\
&& &0.006 & 0.006 & 0.008 & 0.004\\
\cline{3-7}
 &\multirow{3}{*}{Recon} & $\bar{\textit{KG}}$ & 0.025& 0.017& 0.026 & \cellcolor{gray!35}{\textbf{0.028}}\\
&& \multirow{2}{*}{$\bar{\textit{DG}}$ \begin{tabular}[c]{@{}|c@{}}EP\\EO\end{tabular}} &0.015 & 0.014& -0.001& 0.005\\
&&  & 0.040 & 0.036& \underline{0.001} & 0.009\\
\hline
\multirow{6}{*}{\rotatebox{90}{\footnotesize\textbf{Fairness aware}}}&\multirow{3}{*}{L2TFMT} & $\bar{\textit{KG}}$ & 0.024 & -0.005 & -0.022 & -0.020\\
&& \multirow{2}{*}{$\bar{\textit{DG}}$ \begin{tabular}[c]{@{}|c@{}}EP\\EO\end{tabular}} & \underline{0.001} & \underline{0.001} & \underline{-0.002} & \underline{0.0}\\
&&  & \underline{0.002} & \underline{0.003} & \underline{0.001} & \underline{0.003}\\
\cline{3-7}
&\multirow{3}{*}{WB-fair} & $\bar{\textit{KG}}$ & -0.016 & 0.002 & -0.051 &-0.080 \\
&& \multirow{2}{*}{$\bar{\textit{DG}}$ \begin{tabular}[c]{@{}|c@{}}EP\\EO\end{tabular}} & \underline{0.001} & 0.004 & 0.001 & 0.002\\
&&  & \underline{0.002} & 0.006 & 0.003 & 0.007 \\
\hline
\multirow{3}{*}{\rotatebox{90}{\footnotesize\textbf{Our}}}&\multirow{3}{*}{\em \our{}} & $\bar{\textit{KG}}$ & \cellcolor{gray!35}{\textbf{0.036}} &  \underline{0.032}& \cellcolor{gray!35}{\textbf{0.036}}& 0.006\\
&& \multirow{2}{*}{$\bar{\textit{DG}}$ \begin{tabular}[c]{@{}|c@{}}EP\\EO\end{tabular}} & \cellcolor{gray!35}{\textbf{-0.001}}  & \cellcolor{gray!35}{\textbf{0.0}}  & \cellcolor{gray!35}{\textbf{-0.004}}  & \cellcolor{gray!35}{\textbf{-0.001}} \\
&&  & \cellcolor{gray!35}{\textbf{0.0}}  & \cellcolor{gray!35}{\textbf{0.0}} & \cellcolor{gray!35}{\textbf{-0.003}} & \cellcolor{gray!35}{\textbf{0.0}}\\
\hline
\end{tabular}
\end{table}

\noindent\textbf{FairBranch outperforms the competitors on average knowledge and discrimination gain.} 
Table~\ref{tab:exp} presents the $\bar{\textit{KG}}$ and $\bar{\textit{DG}}$ values of various \verb|MTL|s across different data setups. Notably, \our{} achieves the best outcome in 10 instances and the second best in one out of 12 occasions. It is only outperformed by \em{TAG} for $\bar{\textit{KG}}$ on ACS-PUMS 19-21 and by \em{Recon} on CelebA age. Importantly, \our{} consistently achieves positive ($\bar{KG}>0$) average knowledge gain, addressing \nt{}, and non-positive ($\bar{DG}\leq 0$) average discrimination gain, tackling \ut{}. 
Among the competitors, fairness-aware \verb|MTL|s (L2TFMT and WB-fair) handle discrimination gain better than accuracy-based conflict-aware and task-grouping methods, with L2TFMT having a slight edge over WB-fair. However, none of the competitors achieve negative $\bar{DG}$ values, indicating evidence of \ut{} even in fair-\verb|MTL|. 
Accuracy-based conflict-aware MTL methods like PCGrad and Recon excel in achieving positive average knowledge gain across all experiment setups. Task-grouping methods FAFS and TAG perform well in addressing \nt{} on tabular data but exhibit negative knowledge gain on visual data, indicating signs of \nt{}. FairBranch with parameter-based grouping combines the benefits of conflict-awareness and task-grouping, effectively mitigating both negative and bias transfer.

\noindent \textbf{FairBranch tackles \nt{} and \ut{} better than the competitors}. To highlight how \our{} performs against the competitors on \nt{} and \ut{}, we illustrate in Fig.~\ref{fig:overall} the distribution of knowledge gain (see Eq.~\ref{eq.kt}) w.r.t., accuracy, and discrimination gain w.r.t., EP, and EO of each \verb|MTL| over the tasks in each dataset. In each of the boxes, green triangles indicate (`$>0$' for accuracy and `$<0$' for fairness) a positive/unbiased transfer, while red triangles indicate a negative/bias transfer of knowledge. We first note that overall \our{} predominantly exhibits green triangles in accuracy on all data setups, which verifies the achievement of our goal of avoiding \nt{}. On tabular data (Fig~\ref{fig.pums1} and \ref{fig.pums2}) for both the measures EO and EP, our performance is very close (DG(t)$\approx 0$) to that of \verb|STL| in all tasks, thus remaining unaffected from \ut. On visual data (Fig~\ref{fig.celeb1} and \ref{fig.celeb2}), we mostly have unbiased transfer, achieving dense concentration of low green triangles, for both EO and EP. But we still suffer from \ut{} in some of the tasks on both data setups. Interestingly, even the fair-\verb|MTL| methods (L2TFMT and WB-fair) also fail to overcome this challenge, showcasing the difficulty of \ut{} under a large number of tasks. 

\noindent\textbf{Tackling \nt{} on parameter space is advantageous over on output (loss) space}. 
The gradient correction competitors (PCGrad and Recon), although better than \our{} on accuracy by achieving higher positive difference, both fail to tackle \ut{} by consistently producing many red triangles across all data setups. Task-grouping methods (FAFS and TAG) tackle with the \nt{} in tabular data, but collapse when dealing with a large number of tasks in visual data setups. The finding highlights the advantage of focusing on parameter space (like PCGrad and Recon), rather than on actual output space (like FAFS and TAG), and justifies our reason of using parameter similarity to identify task-groups. 
\begin{figure}[hbt!]
    \centering
    \vspace{-0.1in}
         \includegraphics[width=0.7\columnwidth]{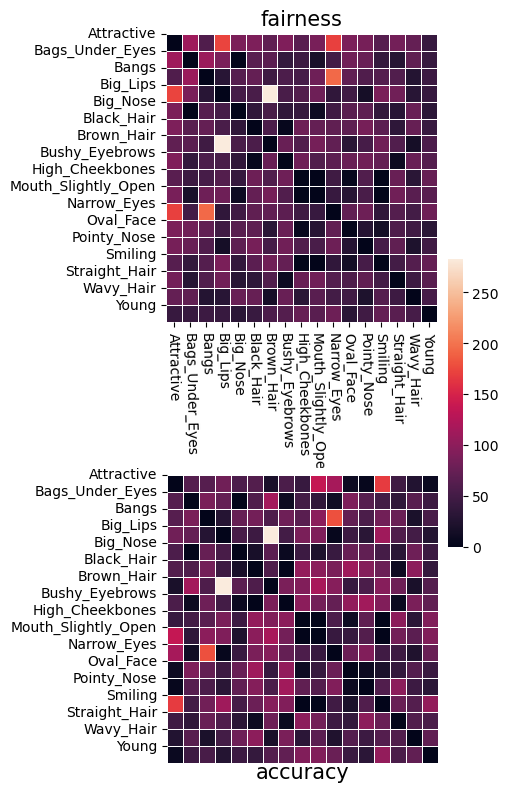}
         \vspace{-.1in}
         \caption{\small Heatmap of Accuracy and Fairness Conflicts on CelebA gen. Brighter colour indicates Higher number of Conflicts.}\label{fig.heat_gen}   
         \vspace{-.1in}
\end{figure}
\subsection{Accuracy and Fairness Conflicts}\label{sec.result_conflict}

In this section, we aim to analyze the reasons behind the errors observed in \our{} in Section~\ref{sec.results}. Despite overcoming the challenge of \nt{} in visual data setups, \our{} still suffers from \ut{} in certain tasks. Our hypothesis suggests that while \our{} effectively resolves accuracy conflicts during training, it struggles to completely eliminate fairness conflicts in certain tasks. To verify this, we plot the distribution of accuracy and fairness conflicts in Fig.~\ref{fig.conflicts_dist}.

\begin{figure}
\vspace{-0.1in}
     \centering
     \includegraphics[width=\columnwidth]{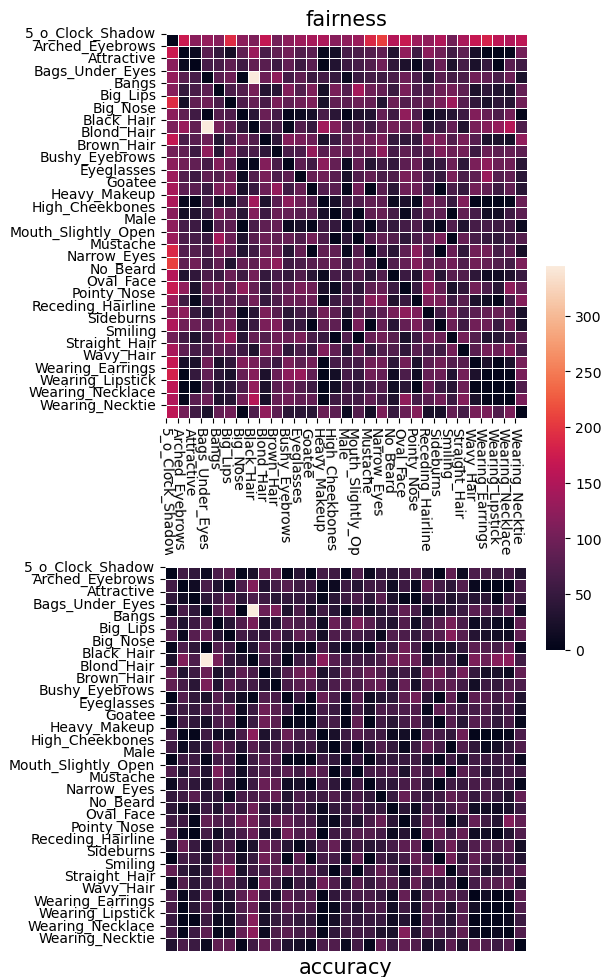}
     \vspace{-.1in}
     \caption{\small Heatmap of Accuracy and Fairness Conflicts on CelebA age. Brighter colour indicates Higher number of Conflicts.}\label{fig.heat_age}
     \vspace{-.1in}
\end{figure}

In both CelebA gen (Fig.\ref{fig.dist_gen}) and CelebA age (Fig.\ref{fig.dist_age}), \our{} reduces both the frequency and severity of conflicts as training progresses. However, towards the end of training, the accuracy conflict boxes are much smaller than the fairness conflict boxes, consistent with our observations in Section~\ref{sec.results}.
We investigate whether conflict occurrence is dominated by a few tasks, given that \ut{} is observed in only a few tasks (cf. Fig.\ref{fig.celeb1}, \ref{fig.celeb2}). Heatmaps of conflicts between tasks accumulated over training epochs are plotted in Fig\ref{fig.heat_gen} and \ref{fig.heat_age}. While no task is free of either accuracy or fairness conflicts, some task pairs exhibit fewer conflicts over multiple epochs and at multiple layer depths during training.

An intriguing observation is that attribute prediction tasks like `Attractive' in Fig.~\ref{fig.heat_gen} and `5\_o\_Clock\_shadow' in Fig.~\ref{fig.heat_age} have fewer accuracy conflicts but more fairness conflicts. These pattern suggests that while such tasks contribute positively to accuracy knowledge transfer, they hinder fairness knowledge transfer for most tasks, highlighting the complex decision-making challenges faced by fair-\verb|MTL|.


\section{Conclusion}
We introduced the study of \ut{} and showed that learning a fair-\verb|MTL| model requires to solve the combined problem of \ut{} to tackle discrimination and \nt{} to tackle accuracy issues. We showed that similar to accuracy conflicts for \nt{}, \ut{} originates from fairness conflicts between task gradients. We proposed \our{}, an in-processing algorithm that tackles the problem at the level of model parameters using parameter similarity-based branching to alleviate \nt{}, and with fairness loss gradients correction for reducing \ut{}. Empirically we show that \our{} outperforms many state-of-the-art \verb|MTL|s for both fairness and accuracy. Our qualitative analysis points out the scalability issues of conflict occurrence in fair-\verb|MTL|, and highlights some open challenges for future work.

\section*{Acknowledgment}
This research work received fund from the European Union under the Horizon Europe MAMMOth project, Grant Agreement ID: 101070285, and also supported by the EU Horizon Europe project STELAR, Grant Agreement ID: 101070122.
\bibliographystyle{IEEEtran}
\bibliography{fb_bib}

\end{document}